\documentclass[11pt]{article}

\usepackage[preprint]{acl}    

\usepackage{times}
\usepackage{latexsym}
\usepackage[T1]{fontenc}
\usepackage[utf8]{inputenc}
\usepackage{microtype}

\usepackage{graphicx}
\usepackage{subcaption}
\usepackage{booktabs}
\usepackage{multirow}
\usepackage{float}     
\usepackage{placeins}

\usepackage{amsmath}
\usepackage{amssymb}
\usepackage{mathtools}
\usepackage{amsthm}

\usepackage{tikz}
\usetikzlibrary{positioning,arrows.meta}

\usepackage{algorithm}
\usepackage{algorithmic}

\usepackage[capitalize,noabbrev]{cleveref}
\usepackage{xcolor}
\usepackage{hyperref}

\theoremstyle{plain}

\theoremstyle{definition}

\theoremstyle{remark}

\usepackage[textsize=tiny]{todonotes}

\newcommand{\RR}{\mathbb{R}}

\newcommand{\dmodel}{d_{\text{model}}}
\newcommand{\da}{d_a}

\title{LongAttnComp: Cross-Family Context Compression \\
       for Long-Context Reasoning}

\author{
  Mengmeng Ji \quad Ravi Shanker Raju \quad Jonathan Lingjie Li \quad Chen Wu \\
  SambaNova Systems, Inc.\\
San Jose, CA, USA \\
  {\small\texttt{\{mengmeng.ji, ravi.raju, jonathan.li, chen.wu\}@sambanovasystems.com}}}
\begin{document}
\maketitle

\begin{abstract}
As real-world applications increasingly require processing
inputs of 100k+ tokens, the gap between context
length and inference efficiency has become a critical
bottleneck. Context compression offers a way to
reduce prefill costs while preserving task accuracy. However,
existing training-free attention-based methods leave
substantial gaps in demanding long-context tasks such as code
reasoning. We present LongAttnComp, a long-context adaptation
of AttnComp that fine-tunes a lightweight cross-attention
scoring layer and introduces token-level chunking, a
token-budget top-$p$ algorithm, positional reordering, and a
format-agnostic query parser. We further design a two-stage fine-tuning recipe for the compressor: Stage 1
builds a general retrieval foundation from NIAH-style data,
and Stage 2 extends it with multi-hop and
reasoning data for broader long-context task coverage. On
InfiniteBench Code-Debug, LongAttnComp matches or exceeds
full-context accuracy, substantially outperforms training-free
baselines, and transfers across four target models from three
families. On LongBench v2, the two-stage
recipe largely closes the Stage 1 gap on
multi-document reasoning while preserving Code-Debug performance. 
\end{abstract}

\section{Introduction}
\label{sec:intro}

Long-context inference with large language models (LLMs) imposes significant
memory and compute costs. As real-world applications
increasingly require processing tens of thousands of tokens — retrieved documents, long
conversations, or extended codebases — the gap between context length and
inference efficiency has become a critical bottleneck. \emph{Context compression}
addresses this concern by filtering or condensing the input context before it reaches
the target model~\citep{jiang2023llmlingua, xu2023recomp}, trading a small
upfront cost for substantial savings in the target model's prefill stage.

\begin{figure}[t]
  \centering
  \begin{tikzpicture}[
    every node/.style={font=\footnotesize, align=center},
    taskbox/.style={rectangle, rounded corners=4pt, draw, thick,
                    fill=gray!8, draw=gray!55,
                    minimum height=9mm, minimum width=62mm,
                    inner sep=4pt, font=\small\bfseries},
    skill/.style={rectangle, rounded corners=4pt, draw, thick,
                  minimum height=18mm, minimum width=28mm, inner sep=4pt},
    retskill/.style={skill, fill=blue!12, draw=blue!55!black},
    reaskill/.style={skill, fill=green!14, draw=green!50!black},
    pluss/.style={font=\large\bfseries, gray!55!black},
    focus/.style={font=\scriptsize\bfseries\itshape, blue!55!black}
  ]
    \node[taskbox] (task) at (0,0)
      {Long-context QA performance};

    \node[retskill] (retr) at (-2.0,-1.9)
      {\textbf{Retrieval}\\[1pt]
       \scriptsize locate query-relevant\\\scriptsize evidence in long input};

    \node[pluss] at (0,-1.9) {+};

    \node[reaskill] (reas) at (2.0,-1.9)
      {\textbf{Reasoning}\\[1pt]
       \scriptsize infer answer from\\\scriptsize retrieved evidence};

    \draw[gray!55, thick] (task.south) -- ++(0,-0.35) -| (retr.north);
    \draw[gray!55, thick] (task.south) -- ++(0,-0.35) -| (reas.north);

  \end{tikzpicture}
  \caption{Long-context task performance decomposes into retrieval
and reasoning~\citep{yang2018hotpotqa}; we focus on the retrieval
bottleneck~\citep{liu2024lostmiddle}, framing compression as a
retrieval problem.}
  \label{fig:motivation}
\end{figure}
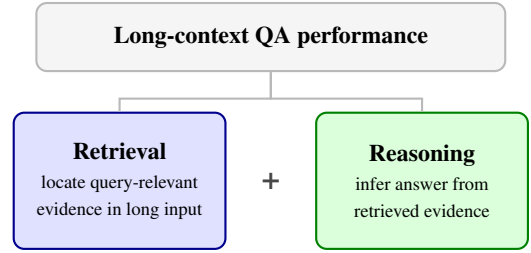

Performance on long-context tasks depends on two factors: reliably retrieving
the query-relevant content, and reasoning correctly over it. This decomposition is illustrated in Figure~\ref{fig:motivation}.
We focus on the retrieval step as the primary bottleneck — effective context
compression is fundamentally a retrieval problem, requiring the compressor to
identify which tokens and segments carry query-relevant information.

Among context compression solutions, Speculative
Prefill~\citep{liu2025specprefill} establishes an attention-based,
training-free, draft-model-driven compression framework with strong
performance across tasks and target model
families~\citep{upasani2026crossfamily}, yet shows a substantial
performance gap on long-context code reasoning.
AttnComp~\citep{luo2025attncomp} takes a fine-tuning approach: a
frozen LLM backbone with a trainable cross-attention layer scores
each document for query relevance. While AttnComp's mechanism
shows promise, its evaluation and training are narrowly
scoped---retrieval-augmented QA at $\sim$12k-token inputs, training
from a single source (HotpotQA), and document-level
scoring---leaving its potential as a general-purpose long-context
compressor untested.

We propose \textbf{LongAttnComp}, a robust, modular long-context
compressor that adapts AttnComp's~\citep{luo2025attncomp}
fine-tuned compression mechanism for the draft-model-driven
framework of Speculative Prefill~\citep{liu2025specprefill},
delivering strong target-model generalization on long-context
retrieval and reasoning tasks. We retain AttnComp's core
mechanism and extend it through four architectural
adaptations and a two-stage training recipe.

\textbf{Architectural adaptations.}
First, we use \textbf{token-level chunking} rather than
document-level scoring, enabling flexible operation on real-world
long-context inputs that often lack natural document boundaries.
Second, we modify AttnComp's score-threshold top-$p$ algorithm with
a \textbf{token-budget variant} that supports both cumulative-score
and budget-only selection modes, giving predictable control over
compression length. Third, we apply \textbf{positional reordering}
to restore selected chunks to their original order before passing
them to the target model. Fourth, we introduce a
\textbf{format-agnostic query parser} for inputs without fixed
query templates.

\textbf{Two-stage finetuning recipe.}
We propose a two-stage fine-tuning recipe that extends
LongAttnComp's effectiveness across diverse long-context tasks.
Stage 1 establishes general query-aligned retrieval capability
on broad NIAH-style data constructed from SQuAD \citep{rajpurkar2016squad} and HotpotQA \citep{yang2018hotpotqa}.
Stage 2 continues training from the Stage 1 checkpoint on a
multi-hop retrieval and reasoning dataset (MuSiQue  \citep{trivedi2022musique},
2WikiMultiHopQA \citep{ho2020wikimultihop}) interleaved with replay from Stage 1 sources.
We further compare two MuSiQue query-construction variants---with and without dataset-provided sub-question decomposition embedded in the query---to characterize how training-time query representation affects downstream task behavior.

On Code-Debug from InfiniteBench \citep{zhang2024infty}, LongAttnComp matches or exceeds
full-context performance, outperforms the
training-free Speculative Prefill baseline, and transfers across
four target models from three families without retraining. On LongBench v2~\citep{bai2024longbench}, Stage 2 training improves on multi-document
reasoning over Stage 1.

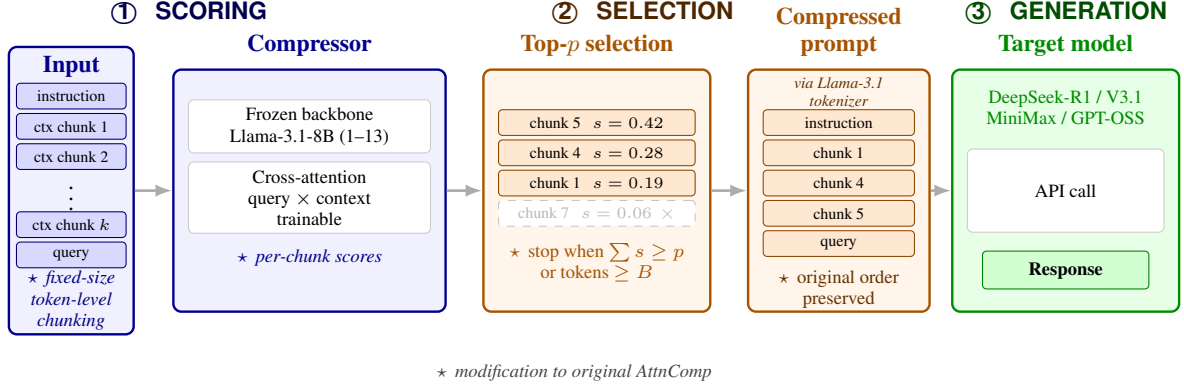
\begin{figure*}[t]
  \centering
  \begin{tikzpicture}[
    font=\footnotesize,
    boxone/.style={rectangle, rounded corners=3pt,
                   draw=blue!55!black, fill=blue!6, line width=0.9pt,
                   align=center, inner sep=4pt},
    boxtwo/.style={rectangle, rounded corners=3pt,
                   draw=orange!65!black, fill=orange!8, line width=0.9pt,
                   align=center, inner sep=4pt},
    boxthree/.style={rectangle, rounded corners=3pt,
                     draw=green!55!black, fill=green!10, line width=0.9pt,
                     align=center, inner sep=4pt},
    innerbox/.style={rectangle, rounded corners=2pt, draw=gray!40, fill=white,
                     align=center, inner sep=3pt, font=\scriptsize},
    chunkone/.style={rectangle, rounded corners=1.5pt,
                     draw=blue!45!black, fill=blue!15,
                     align=center, inner sep=2pt, font=\tiny,
                     minimum width=1.45cm, minimum height=0.34cm},
    chunktwo/.style={rectangle, rounded corners=1.5pt,
                     draw=orange!55!black, fill=orange!18,
                     align=center, inner sep=2pt, font=\tiny,
                     minimum height=0.34cm},
    chunkfaded/.style={rectangle, rounded corners=1.5pt,
                       draw=gray!45, fill=white, dashed, text=gray!55,
                       align=center, inner sep=2pt, font=\tiny,
                       minimum height=0.34cm},
    chunkresp/.style={rectangle, rounded corners=1.5pt,
                      draw=green!50!black, fill=green!18,
                      align=center, inner sep=2pt,
                      font=\scriptsize\bfseries,
                      minimum width=2.2cm, minimum height=0.5cm},
    stagelab/.style={font=\footnotesize\bfseries\sffamily},
    arrow/.style={-{Latex[length=2.2mm]}, draw=gray!65, line width=1.0pt}
  ]

  \node[stagelab, text=blue!30!black]   at (2.4, 4.35)  {\textcircled{\small 1}\ \, SCORING};
  \node[stagelab, text=orange!30!black] at (8.4, 4.35)  {\textcircled{\small 2}\ \, SELECTION};
  \node[stagelab, text=green!30!black]  at (14.0, 4.35) {\textcircled{\small 3}\ \, GENERATION};

  \node[boxone, minimum width=1.65cm, minimum height=3.8cm] (prompt) at (0.85, 2) {};
  \node[font=\footnotesize\bfseries, text=blue!55!black] at (0.85, 3.65) {Input};
  \node[chunkone] at (0.85, 3.25) {instruction};
  \node[chunkone] at (0.85, 2.85) {ctx chunk 1};
  \node[chunkone] at (0.85, 2.45) {ctx chunk 2};
  \node                            at (0.85, 2.00) {$\vdots$};
  \node[chunkone] at (0.85, 1.55) {ctx chunk $k$};
  \node[chunkone] at (0.85, 1.15) {query};
  \node[font=\scriptsize\itshape, align=center,text=blue!50!black] at (0.85, 0.55)
  {$\star$\, fixed-size\\token-level\\chunking};

  \node[boxone, minimum width=3.6cm, minimum height=3.2cm] (compressor) at (4.0, 2) {};
  \node[above=1pt of compressor.north, font=\footnotesize\bfseries, text=blue!55!black] {Compressor};
  \node[innerbox, minimum width=3.2cm] (backbone) at (4.0, 2.85)
    {Frozen backbone\\ \scriptsize Llama-3.1-8B (1--13)};
  \node[innerbox, minimum width=3.2cm, below=3pt of backbone] (xattn)
    {Cross-attention\\ \scriptsize query $\times$ context\\ \scriptsize trainable};
  \node[below=2pt of xattn, font=\scriptsize\itshape, text=blue!55!black]
    {$\star$\, per-chunk scores};

  \node[boxtwo, minimum width=3cm, minimum height=3.2cm] (topp) at (7.8, 2) {};
  \node[above=1pt of topp.north, font=\footnotesize\bfseries, text=orange!65!black] {Top-$p$ selection};
  \node[chunktwo, minimum width=2.6cm] at (7.8, 2.90) {chunk 5 \ \ $s=0.42$};
  \node[chunktwo, minimum width=2.6cm] at (7.8, 2.50) {chunk 4 \ \ $s=0.28$};
  \node[chunktwo, minimum width=2.6cm] at (7.8, 2.10) {chunk 1 \ \ $s=0.19$};
  \node[chunkfaded, minimum width=2.6cm] at (7.8, 1.70) {chunk 7 \ \ $s=0.06$ \ $\times$};
  \node[font=\scriptsize, align=center, text=orange!65!black] at (7.8, 1.05)
    {$\star$\, stop when $\sum s \geq p$\\ or tokens $\geq B$};

  \node[boxtwo, minimum width=2.4cm, minimum height=3.2cm] (compressed) at (11.0, 2) {};
  \node[above=1pt of compressed.north, font=\footnotesize\bfseries,
        text=orange!65!black, align=center] {Compressed\\ prompt};
 \node[font=\tiny\itshape, text=orange!40!black, align=center]
  at (11.0, 3.3) {via Llama-3.1\\ tokenizer};
  \node[chunktwo, minimum width=2.0cm] at (11.0, 2.9) {instruction};
  \node[chunktwo, minimum width=2.0cm] at (11.0, 2.5) {chunk 1};
  \node[chunktwo, minimum width=2.0cm] at (11.0, 2.1) {chunk 4};
  \node[chunktwo, minimum width=2.0cm] at (11.0, 1.7) {chunk 5};
  \node[chunktwo, minimum width=2.0cm] at (11.0, 1.3) {query};
  \node[font=\scriptsize, align=center, text=orange!40!black] at (11.0, 0.70)
    {$\star$\, original order\\ preserved};

  \node[boxthree, minimum width=3.0cm, minimum height=3.2cm] (target) at (14.0, 2) {};
  \node[above=1pt of target.north, font=\footnotesize\bfseries, text=green!55!black] {Target model};
  \node[font=\scriptsize, align=center, text=green!55!black] at (14.0, 3.1)
    {DeepSeek-R1 / V3.1\\ MiniMax / GPT-OSS};
  \node[innerbox, minimum width=2.6cm, minimum height=1.1cm] at (14.0, 2.0) {API call};
  \node[chunkresp] at (14.0, 0.95) {Response};

  \draw[arrow] (prompt.east)     -- (compressor.west);
  \draw[arrow] (compressor.east) -- (topp.west);
  \draw[arrow] (topp.east)       -- (compressed.west);
  \draw[arrow] (compressed.east) -- (target.west);
 \node[font=\scriptsize\itshape, align=center, text=gray!55!black]
  at (7.5, -0.4) {$\star$\, modification to original AttnComp};

  \end{tikzpicture}

  \caption{End-to-end workflow of the LongAttnComp compressor.
\textbf{(1)~Scoring}: a frozen Llama-3.1-8B backbone with a trainable
cross-attention layer produces a relevance score per context chunk.
\textbf{(2)~Selection}: chunks are ranked by score and retained until
cumulative top-$p$ mass or token budget $B$ is reached; selected
token chunks are decoded into a natural-language compressed prompt
by Llama-3.1's tokenizer, with original positional order
preserved.
\textbf{(3)~Inference}: the compressed prompt is sent to the target
model via API.}
  \label{fig:workflow}
\end{figure*}

Our contributions are:
\begin{itemize}

\item \textbf{LongAttnComp}, a trainable draft-model compression
      framework adapted from AttnComp for long-context inference.
      Adaptations include token-level chunking, a token-budget
      top-$p$ algorithm with cumulative-score and budget-only
      selection modes, positional reordering, and a format-agnostic
      query parser (Figure~\ref{fig:workflow}, \S\ref{sec:method}).

\item A \textbf{two-stage fine-tuning recipe} that broadens a LongAttnComp compressor's task coverage and strengthens its multi-hop
reasoning ability, with subq/nosubq query-construction
variants explored as a design lever
(\S\ref{sec:training_design}, \S\ref{sec:data}).
\item \textbf{Empirical findings}: LongAttnComp (i) matches or
      exceeds full-context accuracy on InfiniteBench Code-Debug
      and substantially outperforms training-free baselines,
      (ii) generalizes across four target models from three
      families without retraining, and (iii) the two-stage
      recipe substantially closes the Stage 1 gap on
      LongBench v2 while largely preserving LongAttnComp's long-context code reasoning strength,
      demonstrating that task-sensitivity reflects
      training-data composition rather than an architectural
      limit (\S\ref{sec:results}, \S\ref{sec:discussion}).
\end{itemize}
We plan to release our code and data to facilitate further
research on long-context compression.

\section{Related Work}
\label{sec:related}

\textbf{Context compression.}
Context compression methods fall into two broad categories.
\emph{Abstractive} approaches train auxiliary models to produce
condensed representations of the input~\citep{xu2023recomp,
yoon2024compact}; \emph{extractive} approaches retain a subset of
original tokens or segments using token-level
perplexity~\citep{jiang2023llmlingua, jiang2024longllmlingua} or
embedding-based semantic similarity~\citep{xu2023recomp}. A shared
limitation across both categories is reliance on a predetermined
compression budget that does not adapt to the variable density of
relevant content. Recent adaptive methods address this via
per-sentence relevance classification or threshold-based
scoring~\citep{hwang2024exit, chirkova2025provence}. Our work shares this adaptive philosophy but operates on
\emph{fixed-size token chunks}, using fine-tuned cross-attention
weights as the relevance signal, enabling end-to-end training
within the speculative prefill framework.

\textbf{Attention-based retrieval and speculative inference.}
Speculative Prefill~\citep{liu2025specprefill} demonstrated that
attention weights from a lightweight draft model serve as effective
token-importance signals for training-free compression, with
subsequent work showing transfer across model
families~\citep{upasani2026crossfamily}.
AttnComp~\citep{luo2025attncomp}
introduced an independent attention-based approach: a fine-tuned
cross-attention scoring layer for document-level compression with
adaptive top-$p$ selection, evaluated on short-context
retrieval-augmented QA. Drawing on both lines, LongAttnComp parallels the draft-model
paradigm of speculative decoding~\citep{leviathan2023fast},
fitting within the broader theme of resource-adaptive inference.


\section{Method}
\label{sec:method}

We adapt AttnComp~\citep{luo2025attncomp} for long-context
inference. AttnComp augments the first $L$ layers of a frozen
draft LLM with a trainable cross-attention layer that produces
relevance scores from query--context attention; only
$\sim$0.5\% of parameters are updated. We retain this
architecture and loss formulation
(Appendix~\ref{app:attncomp_review}), extending it in two
ways: architectural adaptations for long-context inputs
(\S\ref{sec:chunking}--\S\ref{sec:parser}) and a two-stage
fine-tuning recipe (\S\ref{sec:training_design}).
Figure~\ref{fig:workflow} shows the full workflow.



\subsection{Token-level Chunking}
\label{sec:chunking}

Whereas the original AttnComp paper uses documents as the unit
for scoring and selection, we operate at token-level chunk
granularity: the input is partitioned into fixed-size token chunks, and the
cross-attention layer scores each chunk independently. This design
is more amenable for real-world long-context inputs, where the input
is often a single long document or stream that cannot be cleanly
partitioned into independent documents. Fixed-size token-level chunking also turns chunk size into a tunable
hyperparameter. Different tasks may benefit from different chunk
sizes, and sweeping chunk sizes across tasks lets us identify the
best size for each task. This, in turn, gives us a clearer
understanding of how chunk granularity affects retrieval and
reasoning in different long-context settings.

\subsection{Modified Top-$p$ Selection}
\label{sec:topp}
We adapt AttnComp's top-$p$ algorithm to operate on chunk-level
scores. The original algorithm stops document selection when
either the cumulative score exceeds $p$ or a document's score
falls below a minimum threshold $\epsilon$. In practice, on long-context tasks, the minimum-score condition
triggers first, causing overly aggressive compression and a
significant performance drop.

We replace the minimum-score threshold with a \textbf{content token
budget} $B$: selection now stops when either the cumulative score
exceeds $p$ or the retained tokens reach $B$
(see Algorithm~\ref{alg:topp}
in Appendix~\ref{app:topp}). This gives direct, predictable control
over compressed context length: with $B=16$k, compressed prompts
consistently approach the budget (15k+ tokens in practice),
avoiding the degenerate under-retention caused by the score
threshold.

\textbf{Positional Reordering.}
After selection, retained token chunks are restored to their
original positional order. The original
AttnComp paper returns the selected documents as an unordered
set; we preserve positional order to maintain discourse coherence.

\textbf{Budget-only selection mode.}
The cumulative-$p$ threshold yields adaptive compression: when a
small subset of chunks captures sufficient query relevance,
selection terminates early and the compressed prompt falls well
below the budget $B$. This works well on tasks where relevant
evidence is concentrated: on RULER's \texttt{niah\_s\_1}, for
example, the compressor reaches 100\% accuracy with prompts
averaging $\sim$2k tokens against a 16k budget, combining
aggressive compression with strong downstream performance. On
tasks where evidence is distributed across many chunks, however,
the same early termination can drop supporting evidence and
reduce accuracy (we characterize this on LongBench v2 in
Appendix~\ref{app:lbv2_selection}). We therefore additionally
support a \textbf{budget-only} mode, which disables the
cumulative-score termination and selects chunks in score order
until the budget $B$ is exhausted. The choice between cumulative
and budget-only selection mode is task-specific, as discussed in
\S\ref{sec:discussion}.
\subsection{Query Parsing}
\label{sec:parser}

AttnComp requires identifying the query span within the concatenated input to
form $X_q$, but the original paper does not discuss query parsing explicitly.
It is understandable given that its standard QA evaluation
uses fixed templates with well-defined query boundaries.

For long-context benchmarks such as InfiniteBench Code-Debug, the prompt
structure is more complex.
We evaluate two parsing strategies. \textbf{Accurate parsing}
precisely extracts the query and instruction spans specific to the
Code-Debug format, providing the compressor with clean,
task-specific query representations. \textbf{Arbitrary parsing}
allocates the last 128 tokens of the input as the query, with the
instruction set identically. Notably, arbitrary parsing incurs
only minor performance degradation relative to accurate parsing,
suggesting LongAttnComp is robust to approximate query
identification. For detailed experimental results, see Appendix ~\ref{app:trainingcodedebug}.

\subsection{Two-Stage Fine-Tuning Recipe}
\label{sec:training_design}

We design the two-stage fine-tuning recipe with two goals:
(i) establish a strong general-purpose retrieval foundation,
and (ii) extend that foundation to harder retrieval patterns
without compromising it.

\textbf{Stage 1: foundation.}
Stage 1 expands AttnComp's training scope in two ways: by
combining single-fact and basic multi-hop retrieval sources to
broaden pattern coverage beyond AttnComp's single-source
training, and by substantially increasing dataset size. Training
hyperparameters follow AttnComp where applicable, with
long-context modifications (cosine LR, dropout, epochs)
selected by ablation (Appendix~\ref{app:ablations}). Evaluating
the resulting checkpoint reveals strong performance on
long-context code reasoning and on single- and basic
multi-needle retrieval, but weak performance on LongBench v2's
natural multi-document reasoning. We hypothesize that the
synthetic, structurally simple nature of NIAH-style training
data limits the compressor's exposure to retrieval patterns
required for multi-document reasoning.

\textbf{Stage 2: extension.}
To test this hypothesis, Stage 2 continues training from the
Stage 1 checkpoint on a dataset targeting harder retrieval
patterns: newly curated multi-hop and naturalistic data
interleaved with replay from Stage 1 sources. Continuing from
the Stage 1 checkpoint avoids re-establishing low-level
retrieval capability from scratch; replay mitigates
catastrophic forgetting of Stage 1 strengths.

\textbf{Sub-question variants.}
Within Stage 2, we additionally probe whether explicit
decomposition of multi-hop questions into their single-hop
constituents during training improves the compressor's
behavior on multi-hop tasks.
MuSiQue's dataset construction~\citep{trivedi2022musique}
composes each multi-hop question from a chain of single-hop
sub-questions and ships these decompositions alongside the
main question. We construct two parallel variants:
\textbf{nosubq} uses the original multi-hop question as the
query verbatim, while \textbf{subq} additionally concatenates
the dataset-provided sub-questions into the query, exposing
the compressor to an explicit reasoning chain. We train and
evaluate both variants (\S\ref{sec:results}), treating
training-time query construction as a deliberate design lever.

Together, the two-stage recipe and its query-construction
variants demonstrate that continued training of a single
cross-attention layer can broaden the compressor's task coverage
while retaining---and in some cases improving---its existing
strengths.

\section{Data}
\label{sec:data}

We build training datasets for both stages using a modified RULER
pipeline~\citep{hsieh2024ruler}: each sample contains 100
candidate documents, a query, and binary relevance labels, with
25\% all-negative samples. Relevance labels come from dataset-provided structural annotations
(SQuAD passage IDs, HotpotQA \texttt{supporting\_facts}, MuSiQue and
2WikiMultiHopQA supporting-paragraph annotations), removing the
need for any LLM-based labeling.

\textbf{Stage~1.}
The Stage~1 training data contains 32,000 examples (16,000
SQuAD~\citep{rajpurkar2016squad}, 0--1 relevant documents per
sample; 16,000 HotpotQA~\citep{yang2018hotpotqa}, 0--2 relevant
documents per sample), with sequences spanning 8k--48k tokens
(data-scale ablation in Appendix~\ref{app:ablations}).

\textbf{Stage~2.}
The Stage~2 training data contains 20,000 samples: 8,000
MuSiQue~\citep{trivedi2022musique}, 4,000
2WikiMultiHopQA~\citep{ho2020wikimultihop}, and 4,000 each from
SQuAD and HotpotQA replay. MuSiQue is constructed in two
variants (subq, nosubq) per \S\ref{sec:training_design}, yielding
two Stage~2 checkpoints. Needle positions across all subsets are
uniformly distributed ($\approx$33\% front/middle/end);
length and position distributions appear in
Appendix~\ref{app:data}.

\section{Experimental Setup}
\label{sec:experiments}

\subsection{Models}

\textbf{Compressor.}
The compressor uses the first $L{=}13$ layers of
\texttt{Llama-3.1-8B-Instruct}~\citep{dubey2024llama3} as a frozen
backbone, with a trainable query-relevant cross-attention layer
appended on top. 

\textbf{Target Models.}
We evaluate compressed prompts on three unrelated target models
accessed via SambaNova Cloud API:
DeepSeek-R1-0528~\citep{deepseek2025r1},
MiniMax-M2.5~\citep{minimax2026m25}, and
GPT-OSS-120B~\citep{openai2025gptoss}. This selection tests both
whether compression retains sufficient information for downstream
answering and whether the compressor generalizes beyond the
\texttt{Llama-3.1-8B-Instruct} family it was trained on. We additionally include \textbf{DeepSeek-V3.1}~\citep{deepseek2025v31}
as a within-family DeepSeek control on Code-Debug and as the
development target for LongBench v2 hyperparameter selection
(Appendix~\ref{app:lbv2_selection}).

\subsection{Compressor Training Details}
\label{subsec: training details}
We hold out 5\% of training data as a validation split and select the
best checkpoint by validation loss. Both stages use the AdamW optimizer
(weight decay $0.01$), batch size 8, gradient accumulation 1, dropout
$0.1$, gradient clipping at $1.0$, and 15 epochs on 8$\times$H200 GPUs,
with a cosine decay schedule and linear warmup.
Stage 1 trains with a learning rate of $2 \times 10^{-4}$. Stage 2 uses
a lower learning rate of $5 \times 10^{-5}$ with warmup over the first
$\sim$5\% of steps.



\subsection{Evaluation Benchmarks}
We center our evaluation of LongAttnComp on \textbf{Code-Debug} from
InfiniteBench~\citep{zhang2024infty}, a multiple-choice
bug-identification benchmark over long code inputs averaging
$\sim$115k tokens with some samples exceeding 200k. We focus on this task for two reasons. First, the original
AttnComp work was evaluated exclusively on Wikipedia-based
QA~\citep{luo2025attncomp}, leaving open whether attention-guided
compression remains effective in substantially longer contexts and
in domains beyond natural language; Code Debug provides a direct
stress test along both axes. Second, the task couples \emph{retrieval} (locating the
relevant buggy region within a long codebase) with
\emph{reasoning} (interpreting code semantics to identify
the correct option), probing what information a compressor
must preserve to support downstream long-context inference.

To assess generalization beyond code, we additionally evaluate on
\textbf{LongBench v2}~\citep{bai2024longbench}, a suite of
long-document tasks requiring multi-hop inference and fine-grained
comprehension across diverse domains, and \textbf{RULER}
~\citep{hsieh2024ruler}, a synthetic suite measuring long-context
utilization across single- and multi-needle retrieval and question
answering.

\subsection{Baselines and Inference Protocol}
\label{subsec:protocol}

We compare LongAttnComp against two baselines:
\textbf{Full context}, the uncompressed prompt sent directly
to the target, and
\textbf{Speculative Prefill}~\citep{liu2025specprefill}, a
training-free attention-based compressor that uses the same
Llama-3.1-8B-Instruct draft model and chunk size 128. The
original AttnComp~\citep{luo2025attncomp} is not included as
a separate baseline since its compressor checkpoint was not
publicly released. For
LongAttnComp, we evaluate three checkpoints: \textbf{Stage 1},
the foundation compressor trained on 32k SQuAD+HotpotQA dataset, and the
two \textbf{Stage 2} variants (\emph{subq}, \emph{nosubq})
that extend Stage 1 with multi-hop reasoning data.

Across all benchmarks, LongAttnComp uses top-$p$=0.95 and a
16k output token budget. Full-context baselines middle-truncate
inputs to fit each target's effective input window (reserved for response), and \texttt{</think>} reasoning
tags are stripped before answer extraction. Per-benchmark
compressor settings differ: chunk size 1024 with query length
128 for Code-Debug, chunk size 256 with query length 256 for
RULER, and chunk size 32 with query length 512 for
LongBench v2.

\section{Results}
\label{sec:results}

\begin{table}[t]
  \centering
  \small
  \begin{tabular}{lc}
    \toprule
    & DeepSeek-R1-0528 \\
    \cmidrule(lr){2-2}
    Method & Accuracy \\
    \midrule
    Full context (120k)        & 74.37 \\
    Speculative Prefill & 62.44 \\
    \midrule
    LongAttnComp (Stage 1)         & 75.38 \\
    LongAttnComp (Stage 2, subq)   & \textbf{76.90} \\
    LongAttnComp (Stage 2, nosubq) & 74.11 \\
    \bottomrule
  \end{tabular}
  \caption{DeepSeek-R1-0528 Accuracy (\%) on Code-Debug of InfiniteBench.}
  \label{tab:codedebug}
\end{table}

We organize our evaluation along two axes. Holding the task fixed at
long-context code reasoning, we ask:
(i) does LongAttnComp match or exceed full-context performance on
InfiniteBench Code-Debug (\S\ref{sec:results_codedebug});
and (ii) does the same compressor, trained with a \texttt{Llama-3.1-8B-Instruct}
draft model, generalize to unrelated target model families (\S\ref{sec:results_crossfamily}).
Holding the compressor fixed and varying tasks, we further ask:
(iii) how does LongAttnComp's effectiveness change across the retrieval
and reasoning demands of LongBench v2 and RULER beyond code
understanding, and what does this reveal about the compressor's
task-type sensitivity (\S\ref{sec:results_lbv2}).

\subsection{Long-Context Code Reasoning}
\label{sec:results_codedebug}

We evaluate LongAttnComp on InfiniteBench Code-Debug with DeepSeek-R1-0528 as the
target model. Table~\ref{tab:codedebug} reports accuracy on
the full test set. Results show that the Stage 1 compressor already exceeds the
full-context baseline and outperforms Speculative
Prefill by 12.9 points. Stage 2 with the \emph{subq} variant pushes
accuracy further to 76.90---the highest in the table---indicating that
adding multi-hop reasoning data in Stage 2 not only preserves Stage 1's
long-context code-reasoning gains but improves on them. 

We adopt arbitrary last-$N$ token query parsing throughout,
removing the deployment dependency on task-specific query
extraction. A direct comparison on the Stage 1 checkpoint
(Appendix~\ref{app:trainingcodedebug}) shows this swap costs at most
1 point on Code-Debug; all main-text LongAttnComp results,
including the Stage 2 numbers in Tables~\ref{tab:crossfamily}--\ref{tab:longbenchv2}, use the arbitrary-query setting.

\subsection{Cross-Family Generalization}
\label{sec:results_crossfamily}
Table~\ref{tab:crossfamily} reports cross-family Code-Debug
results. LongAttnComp Stage 1 matches or closely tracks each
target's full-context accuracy---exceeding DeepSeek-R1-0528 by
1.0 points and trailing DeepSeek-V3.1, MiniMax-M2.5, and
GPT-OSS-120B by under 3 points each---while outperforming
Speculative Prefill by 7--31 points across all four targets.
Stage 2 either improves over or matches Stage 1 on three of
four targets (R1, V3.1, MiniMax), with a small 1--3 point regression on
GPT-OSS.
Because the same Llama-3.1-trained compressor produces all
four blocks without any target-specific fine-tuning or
hyperparameter adjustment, these results support two
conclusions: (i) LongAttnComp transfers cross-family for
long-context code reasoning, and (ii) Stage 2 training is
generally additive on this task, consistent with its goal of
broadening task coverage without compromising Stage 1's
code-reasoning gains.
\begin{table}[t]
  \centering
  \resizebox{\columnwidth}{!}{%
    \begin{tabular}{llcc}
      \toprule
      Target model & Method & Ctx.\ budget & Accuracy \\
      \midrule
      \multirow{5}{*}{DeepSeek-R1-0528}
          & Full context                   & 120k & 74.37 \\
          & Spec Prefill                   & 16k  & 62.44 \\
          & LongAttnComp (Stage 1)         & 16k  & 75.38 \\
          & LongAttnComp (Stage 2, subq)   & 16k  & \textbf{76.90}   \\
          & LongAttnComp (Stage 2, nosubq) & 16k  & 74.11   \\
      \midrule
      \multirow{5}{*}{DeepSeek-V3.1}
          & Full context                   & 128k & 67.51   \\
          & Spec Prefill                   & 16k  & 59.14   \\
          & LongAttnComp (Stage 1)         & 16k  & 65.73   \\
          & LongAttnComp (Stage 2, subq)   & 16k  & 65.48   \\
          & LongAttnComp (Stage 2, nosubq) & 16k  & \textbf{65.99}   \\
      \midrule
      \multirow{5}{*}{MiniMax-M2.5}
          & Full context                   & 152k & 83.76 \\
          & Spec Prefill                   & 16k  & 57.10 \\
          & LongAttnComp (Stage 1)         & 16k  & 81.22 \\
          & LongAttnComp (Stage 2, subq)   & 16k  & \textbf{82.23}   \\
          & LongAttnComp (Stage 2, nosubq) & 16k  & 81.72   \\
      \midrule
      \multirow{5}{*}{GPT-OSS-120B}
          & Full context                   & 120k & 86.00 \\
          & Spec Prefill                   & 16k  & 52.28 \\
          & LongAttnComp (Stage 1)         & 16k  & \textbf{82.99} \\
          & LongAttnComp (Stage 2, subq)   & 16k  & 80.20   \\
          & LongAttnComp (Stage 2, nosubq) & 16k  & 81.73  \\
      \bottomrule
    \end{tabular}%
  }
  \caption{Target-model generalization on Code-Debug.}
  \label{tab:crossfamily}
\end{table}

\begin{table*}[!h]
  \begin{center}
    \begin{small}
      \begin{tabular}{lcccccc}
        \toprule
        
        & \multicolumn{6}{c}{DeepSeek-R1-0528} \\
        \cmidrule(lr){2-7}
        Method & Overall & Easy & Hard & Short & Medium & Long \\
        \midrule
        Full context          & 56.7 & 59.4 & 55.0 & 66.7 & 50.9 & 51.4 \\
        Full context (100k)     & 51.1 & 58.9 & 46.3 & 54.4 & 44.2 & 59.3 \\
        Speculative Prefill                & 46.3 & 49 & 44.7 & 56.7 & 40.5 & 40.7 \\
        LongAttnComp (Stage 1)            & 41.7 & 47.9 & 37.9 & 46.7 & 37.7 & 41.7 \\
        LongAttnComp (Stage 2, subq)          & \textbf{48.9} & \textbf{56.2} & 44.4 & 53.9 & 42.3 & \textbf{53.7} \\
        LongAttnComp (Stage 2, nosubq)          & \textbf{49.7} & 44.2 & \textbf{46.3} & 49.4 & \textbf{49.8} & 50.0 \\
        \bottomrule
      \end{tabular}
    \end{small}
  \end{center}
  \caption{LongBench v2 accuracy (\%), broken down by difficulty
(easy/hard) and input length (short/medium/long). Full context
(100k) reflects a LongBench v2-specific deployment cap of 100k
input tokens; the untruncated full-context row is reproduced
from the LongBench v2 leaderboard~\citep{longbenchv2_leaderboard} for
reference.}
  \label{tab:longbenchv2}
\end{table*}

\subsection{Beyond Code Reasoning: LongBench v2 and RULER}
\label{sec:results_lbv2}

We evaluate LongAttnComp on LongBench v2 to assess performance
on multi-document reasoning.
Table~\ref{tab:longbenchv2} reports accuracy with
DeepSeek-R1-0528, broken down by difficulty and input length.

On LongBench v2, both Speculative Prefill and Stage 1
LongAttnComp underperform the full-context baseline by
substantial margins. We hypothesize that Stage 1's training set
(SQuAD and HotpotQA, NIAH-style) does not cover the
evidence-aggregating reasoning patterns LongBench v2
emphasizes, which is precisely the gap Stage 2 is designed
to close.

We separate hyperparameter selection from final evaluation by
using DeepSeek-V3.1 as a development target and reporting all
main-text numbers on R1 as the held-out target. V3.1 sits
within the same DeepSeek family, making transfer of
inference-time settings plausible, and its non-thinking mode
keeps sweeps tractable. The V3.1 ablation
(Appendix~\ref{app:lbv2_selection}) identifies two impactful
changes from the Code-Debug defaults: token-budget-only
selection (disabling top-$p$ pruning) and a longer parsed
query window ($N{=}512$); both transfer cleanly to R1.

With Stage 2, LongAttnComp recovers 7--12 points over Stage 1
across every breakdown, surpassing Speculative Prefill by
2.6 points (subq) and 3.4 points (nosubq) on Overall
accuracy. The \emph{subq} variant gains the most on Long
inputs (41.7 $\rightarrow$ 53.7); the \emph{nosubq} variant
achieves the highest Overall accuracy (49.7), reaching
within 1.4 points of the 100k-truncated full-context
baseline. While
Stage 2 does not yet match the untruncated full-context
baseline, the consistent improvement over both Stage 1 and
Speculative Prefill validates that broadening the training
set with multi-hop reasoning data transfers to
naturalistic multi-document reasoning.

\textbf{Per-subtask diagnostic on RULER.}
We additionally evaluate on RULER as a complementary
per-subtask probe. LongAttnComp substantially recovers
accuracy on subtasks where the truncated baseline loses to
lost-in-middle effects (e.g., \texttt{niah\_s\_3}:
$57.4 \rightarrow 99.2$) and underperforms on multi-value
and multi-query subtasks where evidence is distributed across
many positions, consistent with the LongBench v2 pattern
above. Stage 2's RULER improvements over Stage 1 are small.
Per-subtask numbers and full analysis are in
Appendix~\ref{app:ruler}.

\section{Discussion}
\label{sec:discussion}


\textbf{Task coverage reflects training data, not architecture.}
The two-stage training and evaluation results show that data
source and distribution have a direct impact on LongAttnComp's
retrieval and reasoning ability. With Stage 1 training,
LongAttnComp performs well on tasks where evidence is clearly
query-aligned and bounded, such as Code-Debug and RULER's
single-needle, multi-key, and QA subtasks. It underperforms when
evidence is indirectly query-aligned or spread across many
locations, such as RULER's multi-value and multi-query NIAH and
LongBench v2's naturalistic multi-document reasoning. This
pattern closely matches Stage 1's training data: synthetic
NIAH-style samples drawn from SQuAD and HotpotQA. We therefore
hypothesize that per-task variation reflects training-data
composition, not a fundamental limit of the method. Stage 2
confirms this. By combining replay of Stage 1 data with newly
curated multi-hop and naturalistic samples (MuSiQue,
2WikiMultiHopQA) that target where Stage 1 is weak, Stage 2
recovers 7 to 12 points across all LongBench v2 breakdowns while
largely preserving Code-Debug performance. Stage 2's improvement
shows that per-task strengths can be extended by fine-tuning on
the right data, without changing the architecture.

\textbf{subq vs.\ nosubq training: mixed evidence.}
Stage 2 is trained in two query-construction variants:
\emph{subq} (multi-hop question plus dataset-provided
sub-question decomposition) and \emph{nosubq} (multi-hop
question only). This tests whether exposing the compressor to
an explicit reasoning chain during training changes its
downstream behavior. The results are mixed. On Code-Debug,
subq wins on two of four targets (R1, MiniMax) and nosubq wins
on the other two (V3.1, GPT-OSS), with per-target differences
under 3 points and no clear pattern across model families. On
LongBench v2, DeepSeek-V3.1 favors subq across every subtask
(Appendix~\ref{app:lbv2_selection}); R1 prefers subq on the
length-bin extremes (Easy, Short, Long) and nosubq on the
mid-range bins (Hard, Medium, Overall).

The V3.1 LongBench v2 result is the most consistent signal in
favor of subq and suggests that explicit sub-question
decomposition during training may help the compressor learn
multi-hop retrieval patterns. Sub-question decomposition is
well-studied as an inference-time strategy for multi-hop
reasoning; applying it as a training-time signal for the
compressor itself is a less-explored variant. Since the clean
win comes from one target on one benchmark, we treat this as
an exploratory observation rather than a firm conclusion, and
leave fuller characterization to future work.

\textbf{Inference-time hyperparameters are task-dependent.}
Two LongAttnComp inference settings interact with the task.
The first is the choice of \textbf{selection mode}. Our modified
top-$p$ algorithm has two termination conditions, cumulative
score and token budget, and adapts the compression length to
the task. On RULER's \texttt{niah\_s\_1}, the top-$p$
threshold is satisfied early and selection stops at $\sim$2k
tokens against a 16k budget, yielding 100\% accuracy with
aggressive compression. On Code-Debug, selection runs close
to the budget ($\sim$15k tokens) because the relevant code
region is larger, again preserving accuracy. On LongBench
v2's naturalistic multi-document reasoning, however,
selection terminates at 6--9k tokens, which is not enough
to retain distributed supporting evidence; switching to
budget-only selection, which disables the top-$p$
termination and fills the full budget, recovers performance across breakdowns
(Appendix~\ref{app:lbv2_selection}). The second knob is
\textbf{chunk size}: 1024 tokens work best for Code-Debug, where
evidence spans full functions; 256 for RULER, where each
needle is short; and 32 for LongBench v2, where supporting
evidence sits in many short spans across documents. Together
these patterns suggest the right inference setting depends
on the task's retrieval and reasoning demands; an adaptive
selection mechanism is a natural direction for deploying
the compressor when task type is unknown.


\textbf{Efficiency.}
Once properly trained, LongAttnComp has a smaller compute and
memory footprint than previous draft-model-based methods such as Speculative
Prefill, which uses the full draft model rather than the first
$L{=}13$ layers. For reference, Speculative Prefill reports a TTFT
reduction from 46s to 2.5s when compressing 128k tokens to 16k
with a Llama-3.1-8B draft model~\citep{upasani2026crossfamily}; since
LongAttnComp's compressor uses only the first 13 of 32 layers of
the same backbone, compression overhead should be roughly
one-third of Speculative Prefill's, at comparable or better
accuracy (Table~\ref{tab:codedebug}).

\section{Conclusion}
\label{sec:conclusion}


We presented LongAttnComp, an effective fine-tuning-based
long-context compression method. The trained compressor acts
as a modular, target-agnostic preprocessing step that
transfers across unrelated target model families without
retraining. Our two-stage training recipe and the resulting
evaluation pattern suggest that with more diverse training
data, the same architecture can extend to more complex
long-context reasoning tasks.

Several future directions follow from this work. First,
expanding the training data beyond Stage 2's MuSiQue and
2WikiMultiHopQA mix, particularly toward more naturalistic
and reasoning-heavy long-context tasks, would further close
the LongBench v2 gap that Stage 2 has partially
addressed. Second, an adaptive mechanism for selecting
inference settings (chunk size and selection mode) would
simplify deployment to inputs whose task type is unknown in
advance. Third, a robust task-agnostic query parser would complete the
deployment story by removing both the task-dependent
query-length choice and the small accuracy cost of arbitrary
last-$N$ parsing ($\sim$1 point on Code-Debug). As a
longer-term direction, fine-tuning the draft model itself to
better align with target model behavior is a natural avenue
we considered but did not pursue here: high-quality
long-context training data remains scarce and generation pipelines for such data
are not openly available.

\section*{Ethics Statement}
All models, datasets, and benchmarks used in this work are
publicly available research artifacts under standard
research-use licenses; our usage is consistent with their
intended research use. The training data is derived from
public QA datasets and is intended for research use only.
We did not collect new data; the source datasets have been
vetted by their original curators and the research community,
and we did not identify personally identifying information
or offensive content in our use. 
\section*{Limitations}
\label{sec:limitations}
We summarize the principal limitations of this work, all of
which are surfaced by the experiments reported in the main
text.

\textbf{Training-data scope.}
Both stages train the compressor on synthetically constructed
NIAH-style data: Stage 1 from SQuAD and HotpotQA, Stage 2
adding MuSiQue and 2WikiMultiHopQA. Despite this broadening,
all training samples are produced by the same synthetic
pipeline that places query-relevant content into
otherwise-irrelevant context. Naturalistic long-context tasks such as LongBench v2 include
reasoning patterns that this synthetic distribution does not
fully capture, which leaves a residual gap to the untruncated
full-context baseline even after Stage 2. Mixing naturally
curated long-context data with the synthetic pipeline is a
necessary next step.

\textbf{Task-dependent hyperparameters.}
LongAttnComp's optimal inference settings vary by task across
three knobs: chunk size, parsed query length, and the choice
between cumulative top-$p$ + budget and budget-only selection
modes. Best chunk size and query length differ across
Code-Debug, RULER, and LongBench v2, and the selection mode
that maximizes accuracy depends on how relevant evidence is
distributed in the input. In settings where the task type is not known in
advance, deploying a single fixed configuration will leave
performance on the table.

\textbf{Query parsing assumption.}
LongAttnComp requires identifying a query span within the
input. We use arbitrary last-$N$ token parsing throughout,
which costs only $\sim$1 point on Code-Debug. This heuristic
introduces two limitations: the optimal $N$ varies by task
(128 for Code-Debug, 256 for RULER, 512 for LongBench v2),
and it is not guaranteed to be sufficient for inputs whose
query is structurally embedded elsewhere in the prompt. A
learned task-agnostic parser would address both.

\textbf{Empirical efficiency measurements.}
We report a rule-of-thumb estimate of compressor overhead
based on the layer-count ratio against a draft-model baseline
(\S\ref{sec:discussion}). We do not provide end-to-end TTFT,
throughput, or memory measurements under controlled hardware;
empirical efficiency characterization is left to follow-up
work.

\textbf{Deployment-side constraints.}
All target-model evaluations use SambaNova cloud
API. Available context budgets, tokenization, and
reasoning-output allocations are determined by the serving
stack rather than the compressor; these constraints
occasionally interact with our evaluation protocol (e.g., the
middle-truncation baseline used in the RULER cross-tokenizer
setting).

\textbf{Single compressor backbone.}
All experiments use Llama-3.1-8B-Instruct as the compressor
backbone. Whether the same training recipe transfers to other
backbone families or scales (smaller draft models for tighter
deployment, larger models for higher headroom) is untested.
\section*{Acknowledgments}
We thank Bo Li for his guidance and support throughout this project, and Taylor Lee for help setting up API endpoints used in our experiments.

\bibliography{emnlp_longattncomp}

@article{luo2025attncomp,
  title   = {{AttnComp}: Attention-Guided Adaptive Context Compression for Retrieval-Augmented Generation},
  author  = {Luo, Lvzhou and Cao, Yixuan and Luo, Ping},
  journal = {arXiv preprint arXiv:2509.17486},
  year    = {2025}
}

@inproceedings{jiang2023llmlingua,
  title     = {{LLMLingua}: Compressing Prompts for Accelerated Inference of Large Language Models},
  author    = {Jiang, Huiqiang and Wu, Qianhui and Lin, Chin-Yew and Yang, Yuqing and Qiu, Lili},
  booktitle = {Proceedings of the 2023 Conference on Empirical Methods in Natural Language Processing},
  pages     = {13358--13376},
  year      = {2023}
}

@article{xu2023recomp,
  title   = {{RECOMP}: Improving Retrieval-Augmented {LMs} with Compression and Selective Augmentation},
  author  = {Xu, Fangyuan and Shi, Weijia and Choi, Eunsol},
  journal = {arXiv preprint arXiv:2310.04408},
  year    = {2023}
}

@inproceedings{leviathan2023fast,
  title     = {Fast Inference from Transformers via Speculative Decoding},
  author    = {Leviathan, Yaniv and Kalman, Matan and Matias, Yossi},
  booktitle = {Proceedings of the 40th International Conference on Machine Learning},
  pages     = {19274--19286},
  year      = {2023}
}

@article{hsieh2024ruler,
  title   = {{RULER}: What's the Real Context Size of Your Long-Context Language Models?},
  author  = {Hsieh, Cheng-Ping and Sun, Simeng and Kriman, Samuel and Agrawal, Shantanu and
             Rekesh, Dima and Fu, Jiaqi and Zhang, Yang and Ginsburg, Boris},
  journal = {arXiv preprint arXiv:2404.06654},
  year    = {2024}
}

@article{zhang2024infty,
  title   = {$\infty${Bench}: Extending Long Context Evaluation Beyond 100K Tokens},
  author  = {Zhang, Xinrong and Chen, Yingfa and Hu, Shengding and Xu, Zihang and Chen, Junhao and
             Hao, Moo Khai and Han, Xu and Thai, Zhen Leng and Wang, Shuo and Liu, Zhiyuan and Sun, Maosong},
  journal = {arXiv preprint arXiv:2402.13718},
  year    = {2024}
}

@article{bai2024longbench,
  title   = {{LongBench} v2: Towards Deeper Understanding and Reasoning on Realistic Long-Context Multitasks},
  author  = {Bai, Yushi and Tu, Shangqing and Zhang, Jiajie and Peng, Hao and Wang, Xiaozhi and
             Lv, Xin and Cao, Shulin and Xu, Jiazheng and Hou, Lei and Dong, Yuxiao and
             Tang, Jie and Li, Juanzi},
  journal = {arXiv preprint arXiv:2412.15204},
  year    = {2024}
}

@article{dubey2024llama3,
  title   = {The {Llama} 3 Herd of Models},
  author  = {Dubey, Abhimanyu and Jauhri, Abhinav and Pandey, Abhinav and others},
  journal = {arXiv preprint arXiv:2407.21783},
  year    = {2024}
}

@inproceedings{rajpurkar2016squad,
  title     = {{SQuAD}: 100,000+ Questions for Machine Comprehension of Text},
  author    = {Rajpurkar, Pranav and Zhang, Jian and Lopyrev, Konstantin and Liang, Percy},
  booktitle = {Proceedings of the 2016 Conference on Empirical Methods in Natural Language Processing},
  pages     = {2383--2392},
  year      = {2016}
}

@inproceedings{liu2025specprefill,
  title     = {Speculative Prefill: Turbocharging {TTFT} with Lightweight and Training-Free Token Importance Estimation},
  author    = {Liu, Jingyu and Chen, Beidi and Zhang, Ce},
  booktitle = {Proceedings of the 42nd International Conference on Machine Learning},
  series    = {Proceedings of Machine Learning Research},
  volume    = {267},
  year      = {2025}
}

@inproceedings{upasani2026crossfamily,
  title     = {Cross-Family Speculative Prefill: Training-Free Long-Context Compression with Small Draft Models},
  author    = {Upasani, Shubhangi and Raju, Ravi Shanker and Li, Bo and Ji, Mengmeng and
               Long, John and Wu, Chen and Thakker, Urmish and Wang, Guangtao},
  booktitle = {International Conference on Learning Representations},
  year      = {2026},
  note      = {arXiv:2603.02631}
}

@inproceedings{yang2018hotpotqa,
  title     = {{HotpotQA}: A Dataset for Diverse, Explainable Multi-hop Question Answering},
  author    = {Yang, Zhilin and Qi, Peng and Zhang, Saizheng and Bengio, Yoshua and Cohen, William W and
               Salakhutdinov, Ruslan and Manning, Christopher D},
  booktitle = {Proceedings of the 2018 Conference on Empirical Methods in Natural Language Processing},
  pages     = {2369--2380},
  year      = {2018}
}

@inproceedings{jiang2024longllmlingua,
  title     = {{LongLLMLingua}: Accelerating and Enhancing {LLMs} in Long Context
               Scenarios via Prompt Compression},
  author    = {Jiang, Huiqiang and Wu, Qianhui and Luo, Xufang and Li, Dongsheng and
               Lin, Chin-Yew and Yang, Yuqing and Qiu, Lili},
  booktitle = {Proceedings of the 62nd Annual Meeting of the Association for
               Computational Linguistics},
  year      = {2024}
}

@article{yoon2024compact,
  title   = {{CompAct}: Compressing Retrieved Documents Actively for Question Answering},
  author  = {Yoon, Chanwoong and Kim, Taewhoo and Hwang, Hyeon and Jeong, Minbyul and
             Kang, Jaewoo},
  journal = {arXiv preprint arXiv:2407.09014},
  year    = {2024}
}

@article{hwang2024exit,
  title   = {{EXIT}: Context-Aware Extractive Compression for Enhancing
             Retrieval-Augmented Generation},
  author  = {Hwang, Taeho and Jeong, Sukmin and Lim, Jeonghwan and Song, Soyeong and
             Park, Jinwoo},
  journal = {arXiv preprint arXiv:2412.12559},
  year    = {2024}
}

@article{chirkova2025provence,
  title   = {Provence: Efficient and Robust Context Pruning for
             Retrieval-Augmented Generation},
  author  = {Chirkova, Nadezhda and Rücklé, Andreas and Gurevych, Iryna and
             Nikoulina, Vassilina},
  journal = {arXiv preprint arXiv:2501.16214},
  year    = {2025}
}

@article{deepseek2025r1,
  title   = {{DeepSeek-R1}: Incentivizing Reasoning Capability in {LLMs}
             via Reinforcement Learning},
  author  = {{DeepSeek-AI}},
  journal = {arXiv preprint arXiv:2501.12948},
  year    = {2025}
}

@misc{minimax2026m25,
  title        = {{MiniMax M2.5}: Built for Real-World Productivity},
  author       = {{MiniMax}},
  year         = {2026},
  howpublished = {\url{https://www.minimax.io/news/minimax-m25}}
}

@article{openai2025gptoss,
  title   = {{gpt-oss-120b} \& {gpt-oss-20b} Model Card},
  author  = {{OpenAI}},
  journal = {arXiv preprint arXiv:2508.10925},
  year    = {2025}
}

@article{liu2024lostmiddle,
  title   = {Lost in the Middle: How Language Models Use Long Contexts},
  author  = {Liu, Nelson F. and Lin, Kevin and Hewitt, John and Paranjape,
             Ashwin and Bevilacqua, Michele and Petroni, Fabio and Liang, Percy},
  journal = {Transactions of the Association for Computational Linguistics},
  volume  = {12},
  pages   = {157--173},
  year    = {2024}
}

@article{trivedi2022musique,
  title     = {MuSiQue: Multihop Questions via Single-hop Question Composition},
  author    = {Trivedi, Harsh and Balasubramanian, Niranjan and Khot, Tushar and Sabharwal, Ashish},
  journal   = {Transactions of the Association for Computational Linguistics},
  volume    = {10},
  pages     = {539--554},
  year      = {2022},
  publisher = {MIT Press},
  doi       = {10.1162/tacl_a_00475},
  url       = {https://aclanthology.org/2022.tacl-1.31/}
}

@inproceedings{ho2020wikimultihop,
  title     = {Constructing a Multi-hop {QA} Dataset for Comprehensive Evaluation of Reasoning Steps},
  author    = {Ho, Xanh and Nguyen, Anh-Khoa Duong and Sugawara, Saku and Aizawa, Akiko},
  booktitle = {Proceedings of the 28th International Conference on Computational Linguistics (COLING)},
  pages     = {6609--6625},
  year      = {2020},
  url       = {https://aclanthology.org/2020.coling-main.580/}
}

@misc{deepseek2025v31,
  title        = {{DeepSeek-V3.1}},
  author       = {{DeepSeek-AI}},
  year         = {2025},
  howpublished = {\url{https://huggingface.co/deepseek-ai/DeepSeek-V3.1}},
  note         = {Hugging Face model release}
}

@misc{longbenchv2_leaderboard,
  title        = {LongBench v2 Leaderboard},
  author       = {{LongBench v2}},
  year         = {2026},
  howpublished = {\url{https://longbench2.github.io/}},
  note         = {Accessed May 2026}
}

\appendix

\section{Background: AttnComp Review}
\label{app:attncomp_review}
We review the AttnComp framework~\citep{luo2025attncomp} that our method
builds upon, covering its architecture, training procedure, and compression
algorithm.

\textbf{Architecture.}
Given an instruction $I$, $k$ retrieved documents
$\mathcal{D} = \{d_1, \ldots, d_k\}$, and a query $q$, the
concatenated input $[I; d_1; \ldots; d_k; q]$ is passed through
the first $L$ frozen layers of a draft LLM, yielding hidden states
$X_c \in \RR^{n \times \dmodel}$ for the context (instruction and
documents) and $X_q \in \RR^{m \times \dmodel}$ for the query.
An additional cross-attention layer---initialized from layer $L{+}1$
of the LLM and the only trainable component---computes
query-context attention weights $A \in \RR^{m \times n}$:
\begin{equation}
 \begin{aligned}
  Q_i = X_q W_i^Q, \quad K_i = X_c W_i^K,  \\
  \quad
  A = \frac{1}{H}\sum_{i=1}^{H} \operatorname{softmax}
      \!\left(\frac{Q_i K_i^\top}{\sqrt{\da}}\right),
 \end{aligned}
  \label{eq:cross_attn}
\end{equation}
where $H$ is the number of attention heads and
$W_i^Q, W_i^K \in \RR^{\dmodel \times \da}$ are per-head projection matrices.
Freezing the first $L$ layers and fine-tuning only the cross-attention layer
updates $\approx 0.5\%$ of total parameters.

\textbf{Training.}
Each training sample contains a query, 100 retrieved documents, and binary
relevance labels $r_i \in \{0,1\}$.
AttnComp trains with two complementary losses.
\textit{Document-level supervision} discriminates relevant from irrelevant
documents via binary cross-entropy:
\begin{equation}
  \mathcal{L}_{\text{doc}} = -\sum_{i=1}^{k}
    \bigl[r_i \log s_{d_i} + (1-r_i)\log(1-s_{d_i})\bigr].
  \label{eq:ldoc}
\end{equation}
\textit{Instruction-level supervision} handles the all-irrelevant case by
directing attention to the instruction when no document is relevant:
\begin{equation}
  \mathcal{L}_{\text{ins}} = -\bigl[r_{\text{ins}} \log s_{\text{ins}}
    + (1-r_{\text{ins}})\log(1-s_{\text{ins}})\bigr],
  \label{eq:lins}
\end{equation}
where $r_{\text{ins}} \triangleq \mathbb{I}\!\left(\sum_{i=1}^{k} r_i = 0\right)$.
The combined loss is $\mathcal{L} = \mathcal{L}_{\text{doc}} + \lambda
\mathcal{L}_{\text{ins}}$.
AttnComp trains on 8k HotpotQA samples (25\% all-negative) using the Adam
optimizer with lr $= 2{\times}10^{-4}$, batch size 8, for 8 epochs with
$\lambda = 0.8$.
Relevance labels are obtained via an automated annotation pipeline.

\paragraph{Top-$p$ compression.}
Each document $d_j$ receives a scalar score $s_{d_j}$ by aggregating rows of
$A$ over its token span; the instruction receives score $s_{\text{ins}}$
similarly.
Documents are sorted descending and a cumulative sum is accumulated starting
from $s_{\text{ins}}$, retaining documents until the sum exceeds threshold $p$
or a score falls below minimum $\epsilon$, yielding $\mathcal{D}^* \subseteq
\mathcal{D}$.
AttnComp uses $p = 0.95$ and $\epsilon = 10^{-2}$.

\section{Our Top-p Algorithm}
\label{app:topp}
Algorithm~\ref{alg:topp} presents our token-budget top-$p$
selection procedure, replacing AttnComp's minimum-score threshold
with a content token budget $B$ that halts selection once the
retained tokens reach $B$ or the cumulative score exceeds $p$.
\begin{algorithm}[H]
  \caption{Token-Budget Top-$p$ Compression}
  \label{alg:topp}
  \begin{algorithmic}
    \STATE \textbf{Input:} Instruction score $s_{\text{ins}}$, document scores
      $\{s_{d_1}, \ldots, s_{d_k}\}$, top-$p$ threshold $p$,
      token budget $B$
    \STATE \textbf{Output:} Compressed document set $\mathcal{D}'$
    \STATE $\{d_{(1)}, \ldots, d_{(k)}\} \leftarrow
      \operatorname{argsort}(\{s_{d_i}\}_{i=1}^{k},\ \text{desc.})$
    \STATE Initialize $\mathit{sum} \leftarrow s_{\text{ins}}$,\
      $\mathcal{D}' \leftarrow \emptyset$,\
      $\mathit{tokens} \leftarrow 0$
    \FOR{$i = 1$ \textbf{to} $k$}
      \IF{$\mathit{sum} \geq p$ \OR
          $\mathit{tokens} + |d_{(i)}| > B$}
        \STATE \textbf{break}
      \ENDIF
      \STATE $\mathit{sum} \leftarrow \mathit{sum} + s_{d_{(i)}}$
      \STATE $\mathit{tokens} \leftarrow \mathit{tokens} + |d_{(i)}|$
      \STATE $\mathcal{D}' \leftarrow \mathcal{D}' \cup \{d_{(i)}\}$
    \ENDFOR
    \STATE \textbf{return} $\mathcal{D}'$
  \end{algorithmic}
\end{algorithm}
\section{Two-Stage Training Data}
\label{app:data}
\begin{figure*}[!h]
  \centering
  \begin{subfigure}[b]{0.23\textwidth}
    \includegraphics[width=\textwidth]{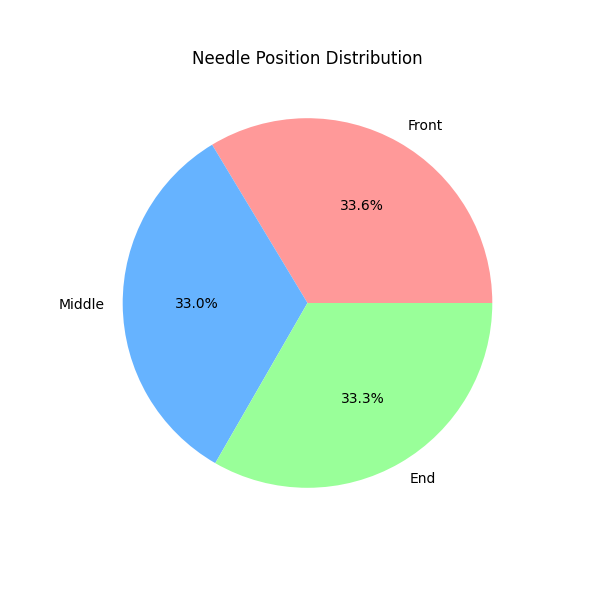}
    \caption{8k-SQuAD}
  \end{subfigure}
  \hfill
  \begin{subfigure}[b]{0.23\textwidth}
    \includegraphics[width=\textwidth]{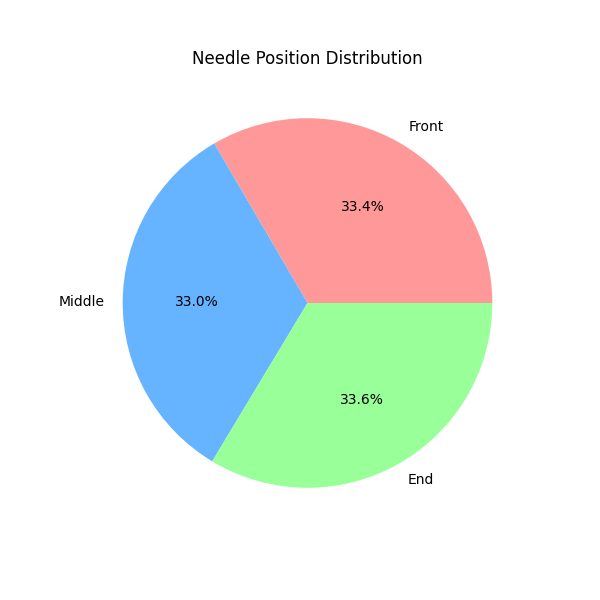}
    \caption{8k-HotpotQA}
  \end{subfigure}
  \hfill
  \begin{subfigure}[b]{0.23\textwidth}
    \includegraphics[width=\textwidth]{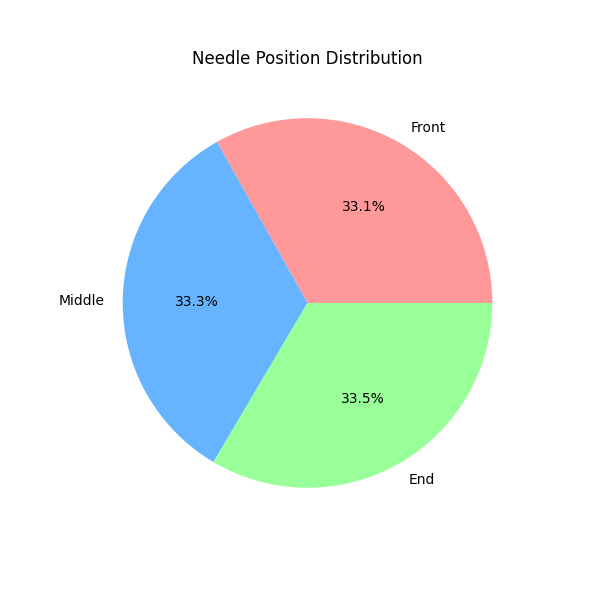}
    \caption{16k-SQuAD}
  \end{subfigure}
  \hfill
  \begin{subfigure}[b]{0.23\textwidth}
    \includegraphics[width=\textwidth]{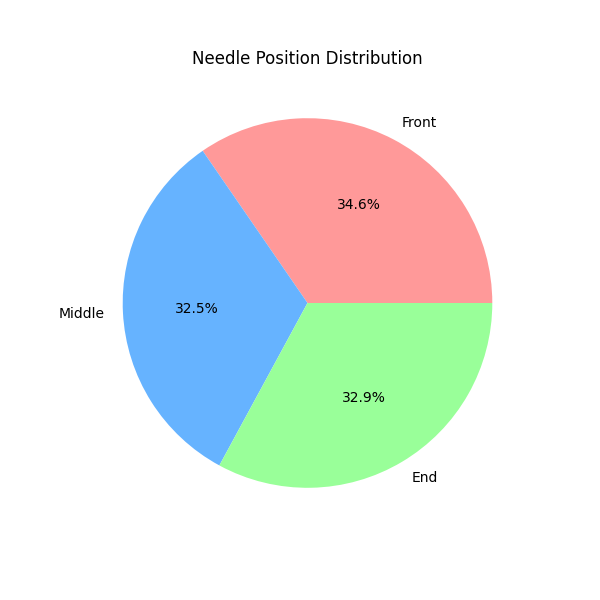}
    \caption{16k-HotpotQA}
  \end{subfigure}
  \caption{Needle position distributions across all training subsets.}
  \label{fig:app_needle}
\end{figure*}

\begin{figure*}[!h]
  \centering
  \begin{subfigure}[b]{0.23\textwidth}
    \includegraphics[width=\textwidth]{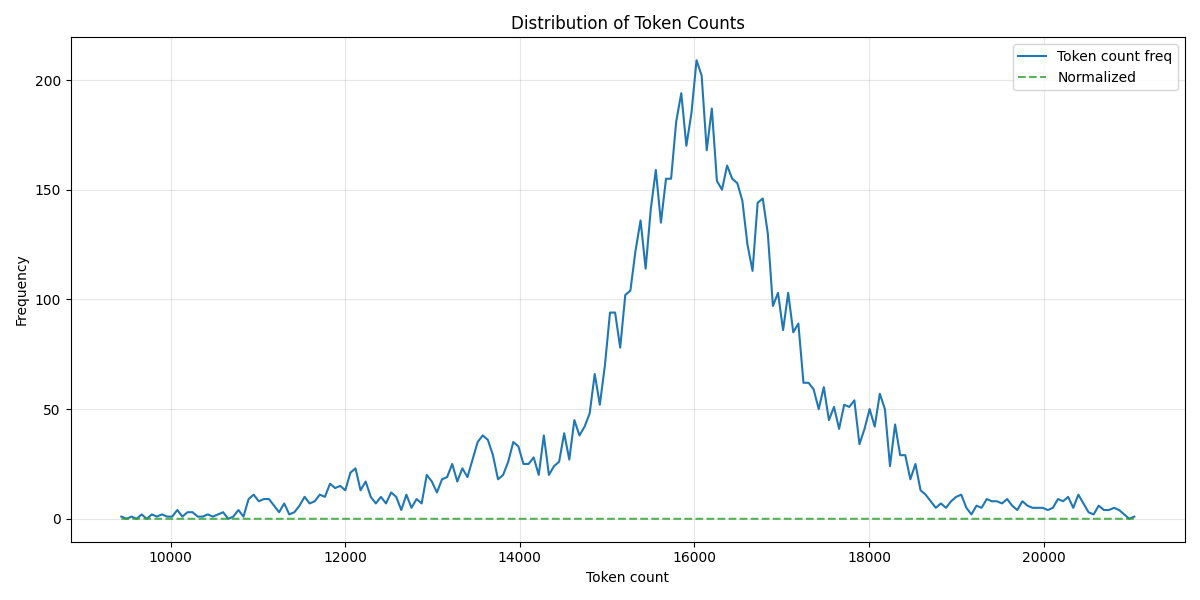}
    \caption{8k-SQuAD}
  \end{subfigure}
  \hfill
  \begin{subfigure}[b]{0.23\textwidth}
    \includegraphics[width=\textwidth]{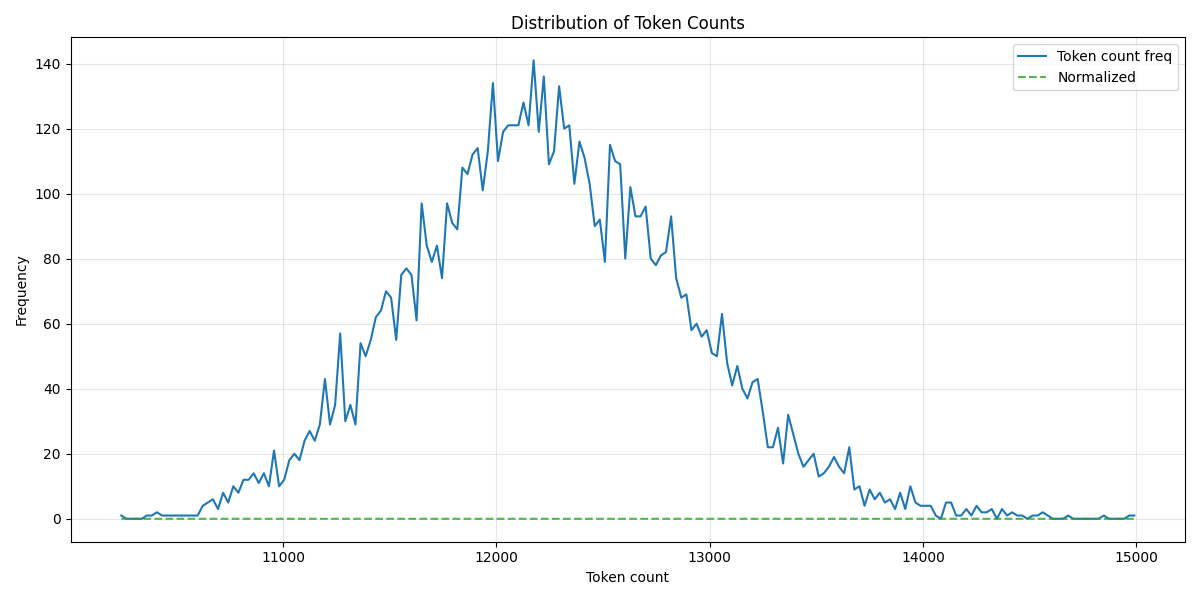}
    \caption{8k-HotpotQA}
  \end{subfigure}
  \hfill
  \begin{subfigure}[b]{0.23\textwidth}
    \includegraphics[width=\textwidth]{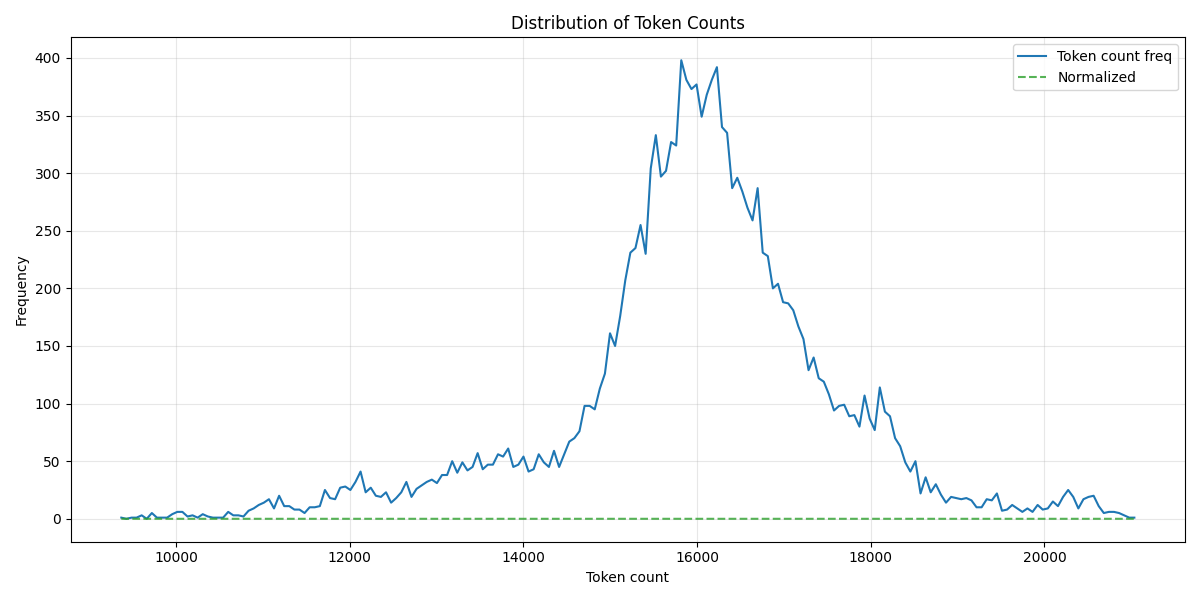}
    \caption{16k-SQuAD}
  \end{subfigure}
  \hfill
  \begin{subfigure}[b]{0.23\textwidth}
    \includegraphics[width=\textwidth]{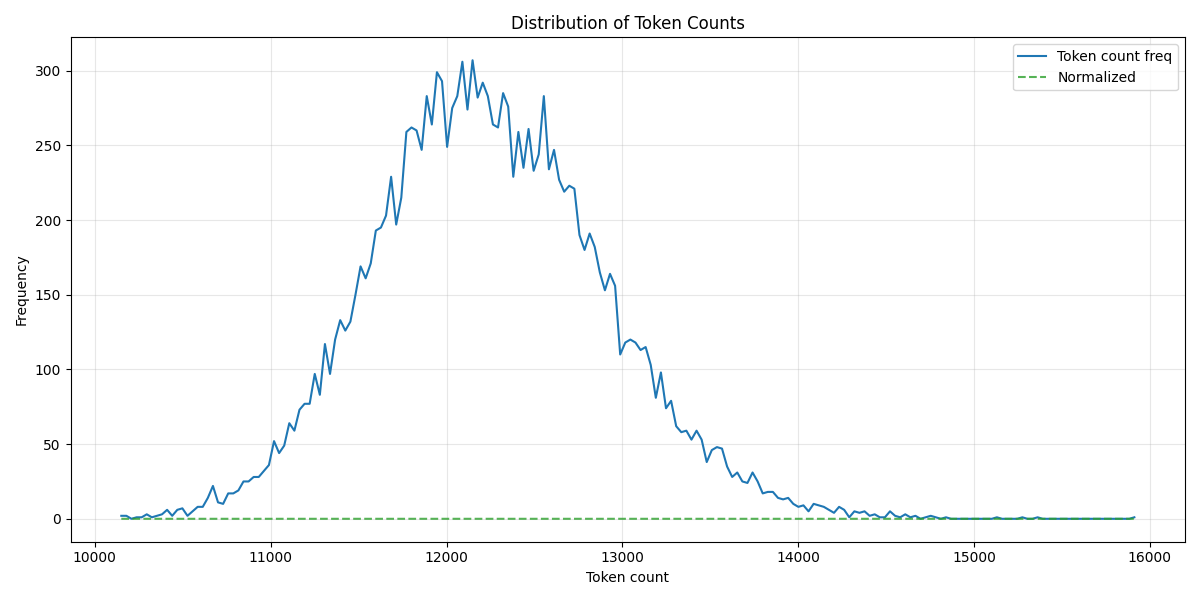}
    \caption{16k-HotpotQA}
  \end{subfigure}
  \caption{Token length distributions across all training subsets.}
  \label{fig:app_token}
\end{figure*}

For \textbf{Stage 1} training data, Table~\ref{tab:data_stats} reports per-subset statistics;
Figures~\ref{fig:app_needle} and~\ref{fig:app_token} show needle positions
and token length distributions for all four training subsets.
All subsets exhibit uniform needle position coverage ($\approx$33\% each
across front, middle, and end), confirming position-agnostic training.

\begin{table}[!h]
  \caption{Summary of Stage 1 training dataset statistics.}
  \label{tab:data_stats}
  \begin{center}
    \begin{small}
      \begin{tabular}{llccc}
        \toprule
        Dataset & Subsets & Samples & Needles & Ctx.\ len \\
        \midrule
        \multirow{2}{*}{16k} & SQuAD    & 8,000  & 0 or 1 & 8k–48k \\
                             & HotpotQA & 8,000  & 0 or 2 & 8k–16k \\
        \midrule
        \multirow{2}{*}{32k} & SQuAD    & 16,000 & 0 or 1 & 8k–48k \\
                             & HotpotQA & 16,000 & 0 or 2 & 8k–48k \\
        \bottomrule
      \end{tabular}
    \end{small}
  \end{center}
  \vskip -0.1in
\end{table}

For \textbf{Stage 2} training data, the replay subsets (SQuAD and
HotpotQA) preserve Stage 1's needle position coverage and
token-length range. The newly added MuSiQue and 2WikiMultiHopQA
subsets differ structurally in both needle count and position
protocol. While Stage 1 samples carry at most 2 relevant documents
(SQuAD: 0--1; HotpotQA: 0--2), Stage 2 samples require 2--4
supporting documents per multi-hop query
(Table~\ref{tab:stage2_stats}; MuSiQue mean 2.33, 2WikiMultiHopQA
mean 2.39). Stage 2 subsets also retain each dataset's natural
needle position distribution rather than enforcing uniform
coverage. The higher needle count substantially raises retrieval
complexity per sample, exposing the compressor to
evidence-aggregation patterns absent from Stage 1.

\begin{table*}[!h]
  \centering
  \small
  \begin{tabular}{lccccc}
    \toprule
    Source & Samples & \multicolumn{3}{c}{Needles per sample} & Tokens (mean) \\
    \cmidrule(lr){3-5}
           &         & 2  & 3  & 4  & \\
    \midrule
    MuSiQue          & 8{,}000 & 73\% & 22\% & 6\%  & 11{,}301 \\
    2WikiMultiHopQA  & 4{,}000 & 80\% & 0\%  & 20\% & 10{,}598 \\
    \bottomrule
  \end{tabular}
  \caption{Stage 2 new-data structural statistics. Needle counts
  reflect the number of supporting documents required per sample;
  token counts are query+context length under the
  Llama-3.1-8B-Instruct tokenizer.}
  \label{tab:stage2_stats}
\end{table*}

\section{Two-Stage Finetuning Recipe}
\label{app:pipeline}
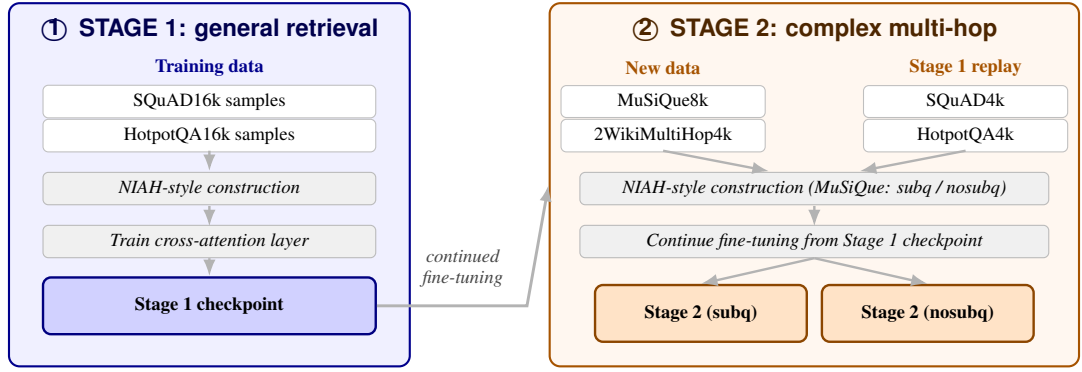
\begin{figure*}[t]
  \centering
  \begin{tikzpicture}[
    font=\footnotesize,
    panelone/.style={rectangle, rounded corners=4pt,
                     draw=blue!55!black, fill=blue!6, line width=0.8pt,
                     inner sep=5pt},
    paneltwo/.style={rectangle, rounded corners=4pt,
                     draw=orange!65!black, fill=orange!6, line width=0.8pt,
                     inner sep=5pt},
    databox/.style={rectangle, rounded corners=2pt,
                    draw=gray!50, fill=white,
                    align=left, inner sep=3pt, font=\scriptsize,
                    minimum height=0.42cm},
    procbox/.style={rectangle, rounded corners=2pt,
                    draw=gray!55, fill=gray!12,
                    align=center, inner sep=3pt, font=\scriptsize\itshape},
    ckptbox/.style={rectangle, rounded corners=3pt, line width=0.9pt,
                    align=center, inner sep=4pt,
                    font=\scriptsize\bfseries,
                    minimum height=0.75cm},
    s1ckpt/.style={ckptbox, draw=blue!50!black, fill=blue!18},
    s2ckpt/.style={ckptbox, draw=orange!55!black, fill=orange!22},
    stagelab/.style={font=\footnotesize\bfseries\sffamily},
    arrow/.style={-{Latex[length=2.2mm]}, draw=gray!60, line width=0.9pt}
  ]

  \node[panelone, minimum width=5.3cm, minimum height=4.8cm] (s1) at (3, 0) {};
  \node[stagelab, text=blue!30!black] at (3, 2.05) {\textcircled{\small 1}\ \,STAGE 1: general retrieval};

  \node[font=\scriptsize\bfseries, text=blue!55!black] at (3, 1.55) {Training data};
  \node[databox, minimum width=4.4cm] at (3, 1.1)  {SQuAD\hfill 16k samples};
  \node[databox, minimum width=4.4cm] at (3, 0.65) {HotpotQA\hfill 16k samples};

  \node[procbox, minimum width=4.4cm] at (3, -0.05) {NIAH-style construction};
  \draw[arrow] (3, 0.45) -- (3, 0.18);

  \node[procbox, minimum width=4.4cm] at (3, -0.75) {Train cross-attention layer};
  \draw[arrow] (3, -0.28) -- (3, -0.55);

  \node[s1ckpt, minimum width=4.4cm] (s1ckptn) at (3, -1.6) {Stage 1 checkpoint};
  \draw[arrow] (3, -0.97) -- (s1ckptn.north);

  \node[paneltwo, minimum width=7.0cm, minimum height=4.8cm] (s2) at (11.0, 0) {};
  \node[stagelab, text=orange!30!black] at (11.0, 2.05) {\textcircled{\small 2}\ \,STAGE 2: complex multi-hop};

  \node[font=\scriptsize\bfseries, text=orange!65!black] at (9.0, 1.55) {New data};
  \node[databox, minimum width=2.7cm] at (9.0, 1.1)  {MuSiQue\hfill 8k};
  \node[databox, minimum width=2.7cm] at (9.0, 0.65) {2WikiMultiHop\hfill 4k};

  \node[font=\scriptsize\bfseries, text=orange!65!black] at (13.0, 1.55) {Stage 1 replay};
  \node[databox, minimum width=2.7cm] at (13.0, 1.1)  {SQuAD\hfill 4k};
  \node[databox, minimum width=2.7cm] at (13.0, 0.65) {HotpotQA\hfill 4k};

  \node[procbox, minimum width=6.2cm] at (11.0, -0.05)
    {NIAH-style construction (MuSiQue: subq / nosubq)};
  \draw[arrow] (9.0, 0.45) -- (10.4, 0.18);
  \draw[arrow] (13.0, 0.45) -- (11.6, 0.18);

  \node[procbox, minimum width=6.2cm] at (11.0, -0.75)
    {Continue fine-tuning from Stage 1 checkpoint};
  \draw[arrow] (11.0, -0.28) -- (11.0, -0.55);

  \node[s2ckpt, minimum width=2.8cm] (s2subq)   at (9.5,  -1.7) {Stage 2 (subq)};
  \node[s2ckpt, minimum width=2.8cm] (s2nosubq) at (12.5, -1.7) {Stage 2 (nosubq)};
  \draw[arrow] (11.0, -0.97) -- (s2subq.north);
  \draw[arrow] (11.0, -0.97) -- (s2nosubq.north);

  \draw[arrow, line width=1.1pt] (s1ckptn.east) -| (7.2, -1.6) -- (s2.west);
  \node[font=\scriptsize\itshape, text=gray!55!black, align=center]
    at (6.35, -1.1) {continued\\fine-tuning};

  \end{tikzpicture}

  \caption{The two-stage fine-tuning recipe. Stage 1 trains the
  cross-attention scoring layer on broad NIAH-style data
  (SQuAD + HotpotQA) to establish general query-aligned retrieval.
  Stage 2 continues fine-tuning from the Stage 1 checkpoint on a
  multi-hop retrieval and reasoning dataset (MuSiQue,
  2WikiMultiHopQA) interleaved with replay from Stage 1 sources.
  MuSiQue admits two variants---with and without sub-question
  decomposition embedded in the query---producing two Stage 2
  checkpoints that we evaluate independently.}
  \label{fig:training}
\end{figure*}

Figure~\ref{fig:training} illustrates the two-stage training
recipe described in \S\ref{sec:training_design} and
\S\ref{sec:data}.


\section{Evaluation Protocols}
\label{app:eval_protocols}

For both Stage 1 and Stage 2, evaluation follows a
tune-then-evaluate procedure: we sweep training and inference
hyperparameters on small held-out subsets to identify the
best configuration, then apply it to the full benchmarks.
For Stage 1, training is tuned on RULER's \texttt{qa\_1} and
\texttt{qa\_2} subtasks (Appendix~\ref{app:ablations}) and
inference settings on a held-out Code-Debug subset. For
Stage 2, inference settings for LongBench v2 are re-tuned on
DeepSeek-V3.1 as a development target before applying to
DeepSeek-R1-0528 (Appendix~\ref{app:lbv2_selection}). Both
stages are then evaluated on Code-Debug, LongBench v2, and
RULER.

\begin{table*}[!t]
  \begin{center}
    \resizebox{\textwidth}{!}{%
      \begin{tabular}{lcccccccccc}
        \toprule
        & \multicolumn{10}{c}{DeepSeek-R1-0528} \\
        \cmidrule(lr){2-11}
        Method
          & niah\_s\_1 & niah\_s\_2 & niah\_s\_3
          & niah\_multik\_1 & niah\_multik\_2 & niah\_multik\_3
          & niah\_multiv & niah\_multiq
          & qa\_1 & qa\_2 \\
        \midrule
        Full context                   & 94.9  & 91.8 & 57.4 & 88.4 & 89.4 & 44.4 & 88.9  & 91.7 & 71.4 & 72.0 \\
        LongAttnComp (Stage 1)         & 100.0 & 99.4 & 98.2 & 92.8 & 73.6 & 80.0 & 64.0  & 81.5 & 89.0 & 83.4 \\
        LongAttnComp (Stage 2, subq)   & 100.0     & 99.2    & 98.6    & 93.6    & 75     & 79.2     & 61.9     & 84    & 88.8    & 82.8    \\
        LongAttnComp (Stage 2, nosubq) & 100.0     & 99.4    & 99.2    & 92.4    & 73.8    & 80.4    & 61.8     & 84.4    & 89.0    & 82.4    \\
        \bottomrule
      \end{tabular}%
    }
  \end{center}
  \caption{Per-subtask diagnostic on RULER. Subtask abbreviations
  follow the RULER conventions~\citep{hsieh2024ruler}.}
  \label{tab:ruler}
\end{table*}
\section{Additional Experiments}
\label{app:ruler}
Table~\ref{tab:ruler} reports per-subtask accuracy on RULER
with DeepSeek-R1-0528 as the target. Three patterns emerge.
First, LongAttnComp substantially recovers accuracy on
subtasks where the truncated baseline loses to lost-in-middle
effects (\texttt{niah\_s\_3}: 57.4 to 99.2;
\texttt{niah\_multik\_3}: 44.4 to 80.4), and matches or
slightly beats the baseline on simpler single-needle and
multi-key tasks. Second, the compressor underperforms the
full-context baseline on multi-value and multi-query
subtasks, where evidence is distributed across many
positions; this spread-evidence weakness relative to full
context mirrors the residual gap on LongBench v2
(\S\ref{sec:results_lbv2}). Third, Stage 2 produces small
improvements over Stage 1 on the multi-needle subtasks
targeted by its multi-hop training data (\texttt{multik\_1},
\texttt{multik\_2}, \texttt{multiq}: 0.8--2.9 points), with
small regressions on multi-value and QA. The Stage 2 gains
on RULER are modest, consistent with the Stage 2 training data
being primarily aimed at LongBench v2-style naturalistic
reasoning rather than RULER's synthetic multi-needle
patterns.

\section{Additional Ablations}
\label{app:ablations}

\begin{table*}[!t]
  \centering
  \resizebox{\textwidth}{!}{%
    \begin{tabular}{lcccccc l}
      \toprule
      & \multicolumn{6}{c}{DeepSeek-V3.1} & \\
      \cmidrule(lr){2-7}
      Method & Overall & Easy & Hard & Short & Medium & Long & Compressed length \\
      \midrule
      Full context                                   & 50.7          & 56.8          & 46.9          & 55.0          & 46.0          & 52.8          & 125k             \\
      \midrule
      Stage 2 (subq), top-$p$ + budget               & 43.5          & 48.4          & 40.5          & 48.3          & 40.9          & 40.7          & 6--9k (adaptive) \\
      Stage 2 (subq), budget-only                    & 45.9          & 51.6          & 42.4          & 48.9          & 43.7          & 45.4          & 16k              \\
      Stage 2 (subq), budget-only, $q$=512           & \textbf{48.9} & \textbf{51.6} & \textbf{47.3} & \textbf{53.3} & \textbf{44.2} & \textbf{50.9} & 16k              \\
      Stage 2 (nosubq), budget-only, $q$=512         & 46.7          & 51.0          & 44.1          & 50.0          & 43.3          & 48.1          & 16k              \\
      \bottomrule
    \end{tabular}%
  }
  \caption{LongBench v2 inference-setting ablation on
  DeepSeek-V3.1 (development target). Two design knobs are swept:
  selection mode (cumulative top-$p$ with budget backup vs.\
  budget-only) and parsed query length ($q{=}128$ default vs.\
  $q{=}512$). The chosen configuration (bolded; Stage 2 subq
  with budget-only selection and $q{=}512$) is applied to
  DeepSeek-R1-0528 as the held-out target in
  Table~\ref{tab:longbenchv2}.}
  \label{tab:app_lbv2_selection}
\end{table*}

\subsection{Compressor Training Ablations}

We conducted ablations throughout development to investigate the
impact of training-data composition and hyperparameters on
LongAttnComp. To select the final training configuration, we
evaluated each candidate checkpoint on the \texttt{qa\_1} and
\texttt{qa\_2} subtasks of RULER. These subtasks probe complementary
skills: \texttt{qa\_1} (SQuAD-derived) tests single-fact retrieval,
while \texttt{qa\_2} (HotpotQA-derived) requires multi-hop retrieval
and reasoning, together covering the retrieval and reasoning axes
central to this work. This choice also matches the QA-centric
evaluation methodology of the original AttnComp paper.

\begin{table}[h]
  
  \begin{center}
    \begin{small}
      \begin{tabular}{lcc}
        \toprule
        Training configuration & qa\_1 & qa\_2 \\
        \midrule
        Llama-3.1, no compression (baseline)        & 71.8 & 43.8 \\
        \midrule
        SQuAD only, const LR, 15 ep.                & 61.4 & 43.0 \\
        HotpotQA only, const LR, 15 ep.             & 26.6 & 51.8 \\
        \midrule
        Combined 16k, const LR, 15 ep.              & 53.8 & 53.4 \\
        Combined 16k, const LR, extended            & 59.4 & 51.8 \\
        \midrule
        Combined 32k, cos LR + dropout, 15 ep.      & \textbf{68.2} & \textbf{58.2} \\
        Combined 32k, cos LR + dropout, 18 ep.      & 65.8 & 57.1 \\
        \bottomrule
      \end{tabular}
    \end{small}
  \end{center}
  \caption{Training-data composition and schedule ablation on RULER
    QA subtasks (\texttt{qa\_1}: SQuAD; \texttt{qa\_2}:
    HotpotQA; Llama-3.1-8B-Instruct as target model, no top-$p$ tuning).
    }
  \label{tab:app_data_ablation}
\end{table}

As shown in Table~\ref{tab:app_data_ablation}, single-source training
produces strong specialization: SQuAD-only and HotpotQA-only
checkpoints each preserve performance on their source task but suffer
significant drops on the other, suggesting that retrieval-biased
training data trades off against reasoning capability and vice versa.
Combining the two sources balances this: even at 16k samples,
performance is balanced across both subtasks, and \texttt{qa\_2}
(the more reasoning-heavy subtask) exceeds the no-compression
baseline. Doubling the dataset to 32k further improves both subtasks.
While this is not a stress test of the cross-attention layer's
training capacity, the fact that the layer absorbs four times the
data used by the original AttnComp~\citep{luo2025attncomp} without
saturating suggests that the lightweight scoring layer can support
substantially larger training sets---a useful design observation
for further scaling. On the hyperparameter axis, we obtain the best results with cosine LR decay, dropout, and 15
training epochs; extending
to 18 epochs produces a small drop, suggesting mild overfitting at
fixed dataset size. These ablations both selected the configuration
used throughout the main text and surfaced general training insights
for single-cross-attention-layer compressors.

\subsection{Stage 1 Training Sweep on Code-Debug}
\label{app:trainingcodedebug}
\begin{table}[!h]
  \begin{center}
    \begin{small}
      \begin{tabular}{lcrr}
        \toprule
        Method                & Chunk & Acc.\ query & Arb.\ query \\
        \midrule
        Full context          &  --   & 74.37       & --          \\
        Speculative Prefill   &  128  & 62.44       & --          \\
        \midrule
        16k, const LR, 15ep   &  128  & 61.42       & --          \\
                              &  256  & 68.02       & --          \\
                              &  512  & 74.37       & --          \\
                              & 1024  & 73.86       & --          \\
        \midrule
        16k, cos LR, 30ep     &  512  & 74.37       & --          \\
                              & 1024  & 75.13       & 75.13       \\
        \midrule
        32k, cos LR, 15ep     &  128  & 56.85       & --          \\
                              &  256  & 72.08       & --          \\
                              &  512  & 72.08       & --          \\
                              & 1024  & \textbf{76.40} & \textbf{75.38} \\
        \bottomrule
      \end{tabular}
    \end{small}
  \end{center}
  \vskip -0.1in
  \caption{Accuracy (\%) on InfiniteBench Code-Debug with DeepSeek-R1-0528.
  All LongAttnComp configurations use top-$p=0.95$ and a 16k token
  budget (compression rate $\approx$83\%). \textbf{Accurate query} uses
  task-specific query extraction; \textbf{Arbitrary query} takes the
  last 128 tokens as the query. Input budget: 120k tokens (8k reserved for output); over-budget inputs
are middle-truncated.\texttt{</think>} tags are stripped before answer extraction.}
  \label{tab:codedebugtrain}
\end{table}

Table~\ref{tab:codedebugtrain} shows the impact of
training-data scale, learning-rate schedule, training-epoch
count, and chunk size on Code-Debug accuracy with
DeepSeek-R1-0528 as the target. Our best Stage 1 configuration
(32k training set, cosine-decay learning rate, 15 epochs,
chunk size 1024, 16k token budget) reaches 76.40\%, exceeding
full context by 2.0 points and Speculative Prefill by 13.9
points while compressing the prompt by 83\%.

Two patterns emerge. First, chunk size has a large impact on
accuracy: on the 32k checkpoint, accuracy increases from
56.85\% at chunk 128 to 76.40\% at chunk 1024. Second,
training schedule and dataset size also contribute: cosine
decay and the 32k training set each improve over their constant-LR
and 16k counterparts, but the 16k cosine checkpoint at 30
epochs already reaches 75.13\%, indicating that longer
training partially compensates for smaller data. Together,
these results indicate that LongAttnComp paired with larger
chunk sizes is particularly effective at retaining
task-relevant information for long-context multiple-choice
reasoning.
\subsection{Top-$p$ Ablation}

Although the original AttnComp paper recommends
$p=0.95$~\citep{luo2025attncomp}, our modified top-$p$ algorithm
and long-context inference setting differ enough that we
re-verified this choice. We sweep $p$ on a small Code-Debug
validation subset, using DeepSeek-R1-0528 as the target model and
the 16k-trained compressor checkpoint. As shown in
Table~\ref{tab:app_topp_ablation}, the sweep is lightweight but
suggestive: $p=0.95$ remains the best choice in our setting, and
we adopt it for all subsequent experiments.
\begin{table}[!h]
  \begin{center}
    \begin{small}
      \begin{tabular}{cc}
        \toprule
        Top-$p$ & Accuracy \\
        \midrule
        0.70          & 50 \\
        0.80          & 60 \\
        0.90          & 50 \\
        \textbf{0.95} & \textbf{90} \\
        \bottomrule
      \end{tabular}
    \end{small}
  \end{center}
  \caption{Top-$p$ threshold ablation. Accuracy (\%) on a 10-sample
  Code-Debug validation subset under our modified token-budget
  top-$p$ algorithm (16k input budget, chunk size 256, 16k training
  set). $p=0.95$ matches the original AttnComp
  default~\citep{luo2025attncomp}.}
    \label{tab:app_topp_ablation}

\end{table}


\subsection{LongBench v2 Inference-Setting Ablation (DeepSeek-V3.1)}
\label{app:lbv2_selection}

We use DeepSeek-V3.1 as a development target to select
inference settings for LongBench v2 before transferring them
to DeepSeek-R1-0528 (\S\ref{sec:results_lbv2}). On V3.1 itself, the
chosen configuration reaches 48.9 Overall, within 2 points
of the full-context baseline (50.7).
Table~\ref{tab:app_lbv2_selection} sweeps two settings:
selection mode (cumulative top-$p$ + budget vs.\ budget-only)
and parsed query length ($q{=}128$ default vs.\ $q{=}512$). Disabling top-$p$ termination and using token-budget-only selection contributes $+$2.4 Overall
points and increasing query length adds another $+$3.0
points, for a total gain of 5.4 over the default. 
The chosen
configuration (Stage 2 subq with budget-only selection and
$q{=}512$) is applied to R1 in the main text.




\end{document}